\documentclass{article}

 \usepackage[preprint]{neurips_2026}

\usepackage[utf8]{inputenc} 
\usepackage[T1]{fontenc}    
\usepackage{hyperref}       
\usepackage{url}            
\usepackage{booktabs}       
\usepackage{amsfonts}       
\usepackage{amsmath}        
\usepackage{amssymb}        
\usepackage{enumitem}       
\usepackage{nicefrac}       
\usepackage{microtype}      
\usepackage[dvipsnames,table]{xcolor}  
\definecolor{ourgrey}{gray}{0.92}      
\usepackage{graphicx}       
\usepackage{algorithm}
\usepackage{algpseudocode}

\usepackage{subcaption}
\usepackage{multirow}       
\usepackage[normalem]{ulem} 
\usepackage{float}          
\usepackage{placeins}       

\title{STRIDE: Training-Free Diversity Guidance via PCA-Directed Feature Perturbation in Single-Step Diffusion Models}

\author{%
  Ankit Yadav \quad Arpit Garg \quad Ta Duc Huy \quad Lingqiao Liu \\
  Australian Institute for Machine Learning, Adelaide University, Australia \\
  \texttt{\{ankit.yadav, arpit.garg, huy.ta, lingqiao.liu\}@adelaide.edu.au} \\
}

\begin{document}

\maketitle

\begin{abstract}
Distilled one-step (T=1) or few-step (T$\leq$4) diffusion models enable real-time image generation but often exhibit reduced sample diversity compared to their multi-step counterparts. In multi-step diffusion, diversity can be introduced through schedules, trajectories, or iterative optimization; however, these mechanisms are unavailable in the few-step or single-step setting, limiting the effectiveness of existing diversity-enhancing methods. A natural alternative is to perturb intermediate features, but naive feature perturbation is often ineffective, either yielding limited diversity gains or degrading generation quality.
We argue that effective diversity injection in few-step models requires perturbations that respect the model's learned feature geometry. Based on this insight, we propose STRIDE, a training-free and optimization-free method that operates in a single forward pass. STRIDE injects spatially coherent (pink) noise into intermediate transformer features, projected onto the principal components of the model's own activations, ensuring that perturbations lie on the learned feature manifold. This design enables controlled variation along meaningful directions in the representation space.
Extensive experiments on FLUX.1-schnell and SD3.5 Turbo across COCO, DrawBench, PartiPrompts, and GenEval show that STRIDE consistently improves diversity while maintaining strong text alignment. In particular, STRIDE reduces intra-batch similarity with minimal impact on CLIP score, and Pareto-dominates existing training-free baselines on the diversity-fidelity frontier. These results highlight that, in the absence of iterative refinement, improving diversity in few-step and one-step diffusion depends not on increasing perturbation strength, but on aligning perturbations with the model's internal representation structure.
\end{abstract}

\section{Introduction}
\label{sec:intro}

 Distilled one-step image generators such as
FLUX.1-schnell~\citep{flux2024} and few-step generators like SD3.5-Turbo~\citep{sd35} have turned
text-to-image generation into a real-time operation, closing
the gap with their multi-step teachers on single-sample
quality while running an order of magnitude faster. A
trade-off of collapsing generation into one or a few
forward passes, however, is reduced sample diversity: given a
fixed prompt and different random seeds, distilled students
often produce visually similar outputs. Gandikota and
Bau~\citep{gandikota2026distilling} measure a twenty-two
percent drop in average pairwise DreamSim relative to an
undistilled teacher, and attribute this to early structural
commitment in the generation process. For applications that
rely on stochastic exploration at a fixed prompt (creative
iteration, dataset synthesis, design variation), few-step
models are efficient but offer limited variability.

Existing approaches to improving diversity largely adapt
techniques developed for multi-step diffusion, such as
conditioning annealing across timesteps~\citep{sadat2023cads},
interval-restricted guidance~\citep{kynkaanniemi2024applying},
trajectory-level optimization~\citep{jalali2025sparke}, or
latent optimization under diversity objectives
(e.g., DPP-based methods~\citep{harrington2025s}). These
methods fundamentally rely on temporal degrees of freedom
(schedules, trajectories, or iterative refinement) that are
absent in the $T{=}1$ or $T{\leq}4$ setting. As a result, when applied to few-step or especially single-step models, they either become ineffective or degenerate to their first iteration, leaving the diversity gap largely unaddressed.

A natural alternative is to introduce stochasticity directly
into intermediate feature representations, treating
transformer blocks as the locus of variation rather than the
noise schedule. Feature perturbation is computationally
efficient, compatible with single-pass inference, and
architecture-agnostic. However, naive feature perturbation
is often ineffective: unstructured noise injected into
intermediate features tends to yield limited diversity gains
or degrade generation quality. This suggests that, in
few-step models, not all perturbations are equally
interpretable as valid variations by the network.

We argue that effective diversity injection requires
perturbations that respect the model's learned feature
structure. In particular, meaningful variation should lie
along directions that are consistent with the model's
internal representation geometry. To this end, we approximate
the local feature manifold using the principal components of
the model's own activations, computed per image. 
This provides a data-dependent basis capturing dominant modes of variation in the current representation. Constraining perturbations to this subspace lets noise be interpreted as plausible variation rather than arbitrary corruption.

Based on this insight, we introduce \textbf{STRIDE}, a
training-free method that injects structured noise into
intermediate transformer features during a single forward
pass. STRIDE combines three key design choices: (i)
spatially coherent ($1/f$-spectrum) noise to preserve
structured variation across tokens, (ii) projection onto
the top $K$-principal components of feature activations to
approximate the learned feature manifold, and (iii)
injection at early, structure-forming layers where global
composition is determined. These components work jointly to
enable controlled diversity while maintaining fidelity.

Our method is implemented via a single forward hook and
requires no training, backpropagation, or iterative
optimization, with a single scalar controlling perturbation
strength. To our knowledge, STRIDE is among the first methods
that simultaneously operate in a training-free,
inference-only, and single-forward-pass regime for diversity
enhancement in distilled diffusion models. 
On FLUX.1-schnell and SD3.5-Turbo across DrawBench, PartiPrompts, COCO, and GenEval, STRIDE consistently improves diversity while preserving text alignment, Pareto-dominating training-free baselines on the diversity-fidelity frontier (Figure~\ref{fig:pareto_main}). Full code will be released upon acceptance (Supp. Sec B.6)

\begin{figure*}[t]
\centering
\begin{subfigure}[t]{0.55\textwidth}
    \centering
    \includegraphics[width=\linewidth]{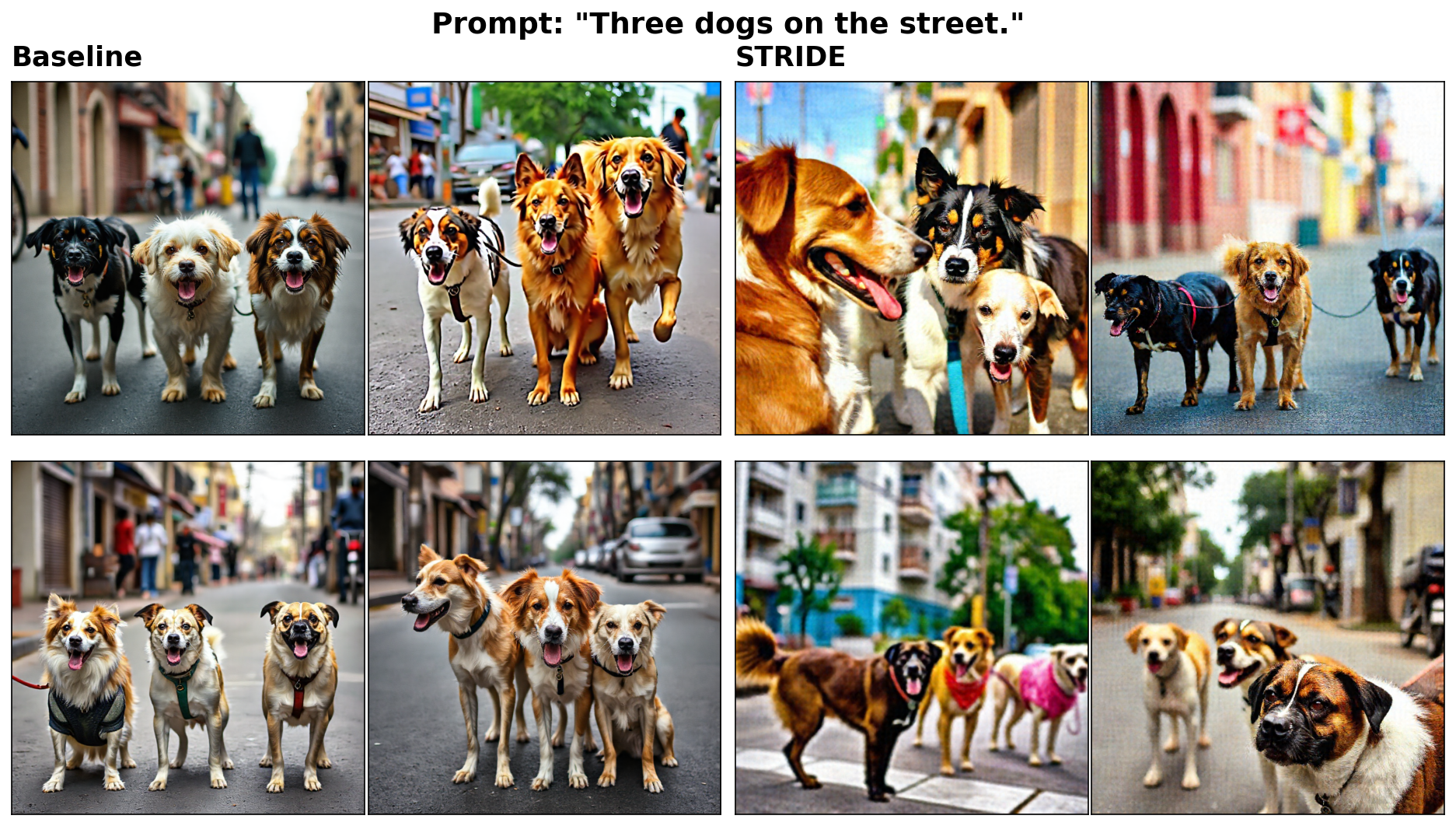}
    \subcaption{}
    \label{fig:qualitative_teaser}
\end{subfigure}
\hfill
\begin{subfigure}[t]{0.43\textwidth}
    \centering
    \includegraphics[width=\linewidth]{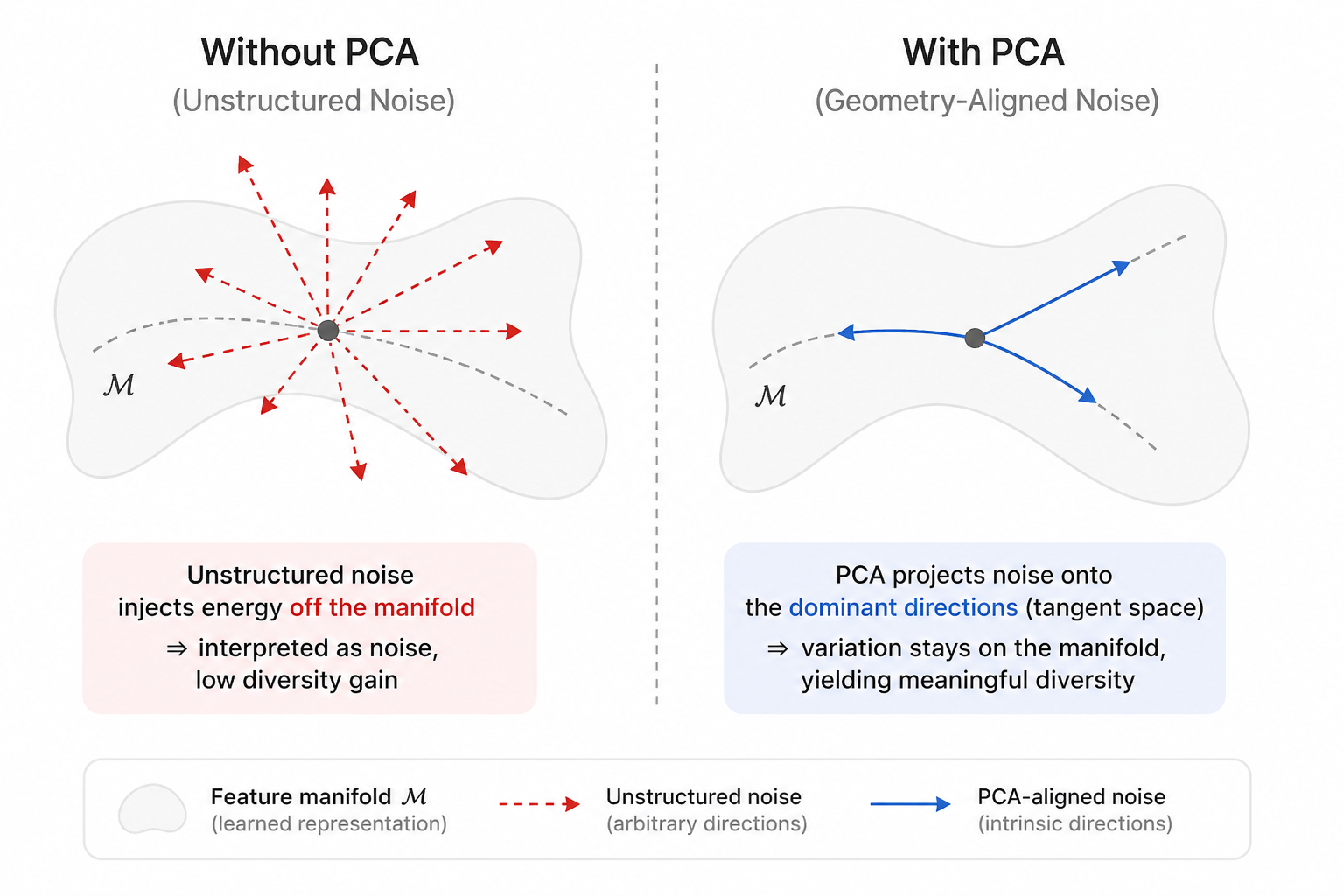}
    \subcaption{}
    \label{fig:pca_manifold}
\end{subfigure}
\caption{\textbf{STRIDE motivation.} 
\textbf{(a)} STRIDE vs.\ random-seed baseline on FLUX.1-schnell: 
each block shows four samples from one DrawBench prompt; STRIDE produces diverse poses and styles. 
\textbf{(b)} Unstructured noise pushes features off-manifold and is treated as 
corruption, while PCA-directed perturbation keeps noise on the 
learned manifold $\mathcal{M}$.}
\label{fig:teaser}
\end{figure*}

\section{Related Work}
\label{sec:related_work}

\noindent \textbf{Mode collapse in distilled one-step generators.}
Adversarial, consistency, and score-distillation schedulers
\citep{sauer2024adversarial, song2023consistency, yin2024dmd}
collapse multi-step denoising into one forward pass but
reduce per-prompt sample diversity. Gandikota and Bau
\citep{gandikota2026distilling} report a twenty-two percent
drop in pairwise DreamSim (0.337 to 0.264) on SDXL-DMD2 and
localize the cause to a first-step structural commitment in
the distilled student; their teacher-for-step-one remedy is
unavailable at $T{=}1$. Training-time counterparts such as
DP-DMD \citep{song2026dpdmd} replace the distillation
objective itself. To our knowledge, STRIDE is the first
diversity method that is simultaneously training-free,
inference-only, and single-forward-pass at $T{=}1$: every
prior method drops at least one of these three axes.

\noindent \textbf{Inference-time remedies that assume $T>1$.}
CADS \citep{sadat2023cads} anneals conditioning across
timesteps; its schedule is undefined at $T{=}1$.
Kynk\"a\"anniemi et al.\ \citep{kynkaanniemi2024applying}
restrict guidance to a middle noise-level interval and show
early denoising steps set diversity, a finding STRIDE
operationalizes when the interval collapses to the only step
available. SPARKE \citep{jalali2025sparke} backpropagates a
kernel-entropy repulsion reward through the denoising
trajectory, which is structurally inapplicable at $T{=}1$.
TPSO \citep{meng2025training} iteratively searches text-embedding
offsets under a dissimilarity objective. Harrington et al.\
\citep{harrington2025s} replace the initial latent with
pink noise and optimize the batch via a Determinantal Point
Process against an HPSv2 quality kernel, requiring roughly
fifteen forward passes per batch. STRIDE retains Harrington's
spectral prior (pink beats white) but forgoes the DPP loop
and its quality kernel for a single-pass PCA projection onto
the learned feature manifold. Particle Guidance
\citep{corso2023particle} and SGI \citep{parmar2025scaling}
enforce batch-level diversity via joint potentials or
post-hoc selection and are composable with ours.

\noindent \textbf{Feature-level perturbation at inference.}
Self- and perturbed-attention guidance
\citep{hong2023improving, ahn2024self, hong2024seg} target fidelity
and reduce diversity as a side effect. Token Perturbation
Guidance (TPG \cite{rajabi2025token}) is the closest
neighbor: it shuffles DiT tokens at selected layers but
requires two forward passes per timestep and applies an
unstructured permutation agnostic to the model's feature
geometry. 

STRIDE uses a single pass and projects the perturbation onto the current block's principal directions, so the injected energy lies on the learned feature manifold rather than its orthogonal complement.

\noindent \textbf{Principal components of diffusion features.}
Concept Sliders \citep{gandikota2024concept}, SliderSpace
\citep{gandikota2025sliderspace}, and h-space analyses
\citep{kwon2023h} identify semantically meaningful directions
in diffusion feature spaces for targeted attribute editing;
each principal component there serves as a controllable axis
to move along. STRIDE uses the same geometric objective (principal components of a block's activations) for a
complementary purpose: rather than move along a single
learned direction, it injects spatially coherent pink noise inside
the entire 
principal subspace, so each generation takes
an independent structural step that preserves on-manifold
plausibility but leaves its content unconstrained. Recent
low-rank analyses of diffusion-transformer activations
\citep{wang2025massive} are consistent with this subspace
view.

\begin{table*}[t]
\centering
\caption{%
  \textbf{Main results on COCO} (2000 prompts $\times$ 10 images).
  STRIDE achieves the best diversity-quality tradeoff among 
  single-pass training-free methods on both FLUX.1-schnell and 
  SD3.5 Turbo. DDC reaches a comparable Pareto point on SD3.5 
  Turbo at the cost of doubled inference-time VRAM. KID values 
  are $\times 10^{-3}$. $^{\dagger}$CADS collapses on SD3.5 
  Turbo. Per-prompt standard deviations and significance analysis 
  in Tables~11 and 12. \textbf{Bold} indicates the better result between STRIDE and its no-PCA ablation.%
}
\label{tab:main_coco}
\small
\setlength{\tabcolsep}{3pt}
\renewcommand{\arraystretch}{1.2}
\begin{tabular}{l ccccc ccccc}
\toprule
& \multicolumn{5}{c}{\textbf{FLUX.1-schnell}~1-Step} & \multicolumn{5}{c}{\textbf{SD3.5 Turbo}~4-Step} \\
\cmidrule(lr){2-6}\cmidrule(lr){7-11}
\textbf{Method}
  & \textbf{InBSim}$\downarrow$ & \textbf{VD/p}$\uparrow$ & \textbf{CLIP}$\uparrow$ & \textbf{HPS}$\uparrow$ & \textbf{KID}$\downarrow$
  & \textbf{InBSim}$\downarrow$ & \textbf{VD/p}$\uparrow$ & \textbf{CLIP}$\uparrow$ & \textbf{HPS}$\uparrow$ & \textbf{KID}$\downarrow$ \\
\midrule
Baseline        & 0.738 & 2.794 & 26.69 & 0.293 & 6.27          & 0.771 & 2.527 & 26.77 & 0.290 & 9.62 \\
CADS            & 0.659 & 3.433 & 26.44 & 0.291 & 6.75          & 0.391$^\dagger$ & 5.888$^\dagger$ & 22.56$^\dagger$ & 0.263$^\dagger$ & 17.54$^\dagger$ \\
Input Noise     & 0.661 & 3.443 & 26.11 & 0.288 & 9.03          & 0.749 & 2.703 & 26.72 & 0.291 & 9.14 \\
DDC             & --    & --    & --    & --    & --            & 0.679 & 3.301 & 26.70 & 0.276 & 7.12 \\
SPELL           & --    & --    & --    & --    & --            & 0.603 & 3.971 & 24.44 & 0.247 & 39.91 \\
\midrule
STRIDE No-PCA       & 0.762 & 2.603 &\textbf{ 28.05} & 0.276 & 8.09          & 0.766 & 2.573 & \textbf{26.82} & 0.289 & 9.41 \\
\rowcolor{ourgrey}
\textbf{STRIDE (Ours)} &\textbf{ 0.683} & \textbf{3.268} & 27.66 & \textbf{0.283} & \textbf{5.74 }         & \textbf{0.725} & \textbf{2.905} & 26.69 & \textbf{0.289} & \textbf{8.68} \\
\bottomrule
\end{tabular}
\end{table*}

 
\section{Method}
\label{sec:method}

We present \textbf{STRIDE} (\textbf{S}patial \textbf{T}ransformer
\textbf{R}epresentation-space \textbf{I}njection for \textbf{D}iversity
\textbf{E}nhancement), a training-free method for increasing output
diversity in single or few-step diffusion transformers without modifying weights or requiring iterative optimization. STRIDE injects structured noise into intermediate transformer representations during a single forward pass.

The key idea is to align perturbations with the model's internal
representation structure (Figure~\ref{fig:pca_manifold}). To this end, STRIDE projects noise onto the
principal components of the model's feature activations, computed
per image. This projection provides a data-dependent approximation of
the local feature manifold, capturing dominant directions of variation
in the representation space. By constraining perturbations to this
subspace, STRIDE enables controlled diversity while preserving semantic
consistency.

\subsection{Motivation}
\label{sec:motivation}

A natural approach to recovering diversity in distilled models is to perturb the initial latent~\cite{harrington2025s}. To evaluate this strategy, we conduct controlled experiments on FLUX.1-schnell by injecting pink noise\footnote{As suggested by \cite{harrington2025s} and confirmed in our study, injecting pink noise can achieve better diversity.} at the input latent and measuring its effect on output diversity (Table~\ref{tab:main_coco}). We find this to be insufficient at $T{=}1$, since input-level pink noise yields only modest diversity gains on FLUX.1-schnell (Table~\ref{tab:main_coco}). This suggests that, in single-step models, much of the stochastic variation introduced at the input is suppressed by the deterministic mapping of the network. An alternative is to perturb intermediate features, which may offer a more direct handle on the learned representation. However, naive feature perturbation proves counterproductive: unstructured noise disrupts the learned representation and increases InBSim above baseline.

The key insight of this paper is that not all directions in feature space correspond to meaningful variation. The model’s representations concentrate along a low-dimensional manifold shaped by dominant modes of variation, and perturbations that deviate from this manifold are interpreted as noise rather than diversity. Therefore, useful variation must be introduced along these intrinsic directions, as illustrated in Figure \ref{fig:pca_manifold}. This motivates the design of structured, feature-aware perturbations that explicitly respect the geometry of the learned representation space.

\subsection{Feature-Level PCA-Directed Noise Injection}
\label{sec:pca_injection}

Building on the insight that meaningful variation must align with the intrinsic geometry of the feature space, we construct perturbations that are explicitly constrained to lie along the dominant modes of the model’s representations. Concretely, we achieve this via a PCA-directed noise injection mechanism applied to intermediate transformer features.

Let $\mathbf{h} \in \mathbb{R}^{B \times N \times D}$ denote the
hidden states at a target transformer block, where $B$ is the batch size,
$N = H \times W$ is the number of spatial tokens, and $D$ is the
feature dimension. STRIDE modifies $\mathbf{h}$ via a forward hook
applied after each target block:
\begin{equation}
    \mathbf{h}' = \mathbf{h} + \alpha \cdot \mathbf{d}
    \label{eq:stride_injection}
\end{equation}
where $\alpha$ is a scalar perturbation strength and $\mathbf{d}$ is a 
manifold-aligned structured perturbation.

To construct geometry-aligned perturbations, we must first identify the dominant modes of variation in the feature space. However, directly applying PCA to token-wise features treats each spatial location independently and fails to capture coherent spatial structures. 
We therefore group spatially adjacent tokens into patches, enabling the subsequent analysis to reflect structured, spatial patterns.

\noindent \textbf{Step 1: Spatial Patchification.}
We reshape the spatial grid of tokens into $P \times P$ patches with 
stride $S$, concatenating the features within each patch:
\begin{equation}
    \mathbf{h}_\text{2D} \in \mathbb{R}^{H \times W \times D}
    \;\longrightarrow\;
    \mathbf{H}_\text{patches} \in \mathbb{R}^{M \times (P^2 D)}
    \label{eq:patchify}
\end{equation}
where $M = \lfloor (H{-}P)/S + 1 \rfloor \times \lfloor (W{-}P)/S + 1 \rfloor$ 
is the number of patches and $P^2 D$ is the concatenated feature 
dimension. 
This couples adjacent tokens so that PCA captures spatially coherent modes of variation (e.g., layout and composition).

\noindent \textbf{Step 2: PCA.}
We compute truncated PCA on the centered patch features of the
\textit{current} image:
\begin{equation}
    \bar{\mathbf{H}} = \mathbf{H}_\text{patches} -
    \boldsymbol{\mu}, \qquad
    \bar{\mathbf{H}} \approx \mathbf{U} \mathbf{S} \mathbf{V}^\top
    \label{eq:pca}
\end{equation}
where $\mathbf{V} \in \mathbb{R}^{P^2 D \times K}$ contains the
top-$K$ right singular vectors (principal directions) and
$\mathbf{S} = \text{diag}(s_1, \ldots, s_K)$ contains the
corresponding singular values. We compute this via randomized SVD~\cite{halko2011finding} (\texttt{torch.pca\_lowrank}, $q{=}K$, niter$=$2).

The PCA is computed \textbf{online} from each image's own
activations to ensure that the principal directions are \textit{prompt-specific}: the spatial PCA basis for ``a red car'' differs from that of ``an astronaut on the moon,'' capturing the content-specific modes of variation available to the model (Supp. Table~6). Crucially, these directions define the local tangent space of the model’s representation manifold, i.e., the directions along which variation is semantically meaningful.

\noindent \textbf{Step 3: Noise Generation and Projection.}
We generate spatially correlated noise via a 2D FFT-based filter 
$\mathcal{F}$ that biases the perturbation toward low spatial frequencies as discussed in \citep{harrington2025s}:
\begin{equation}
    \boldsymbol{\epsilon}_\text{pink} =
    \mathcal{F}^{-1}\!\left[
        \frac{1}{(1 + |\mathbf{f}|)^{f_\alpha}}
        \cdot \mathcal{F}[\boldsymbol{\epsilon}_\text{white}]
    \right]
    \label{eq:pink_noise}
\end{equation}
where $\boldsymbol{\epsilon}_\text{white} \sim \mathcal{N}(0, 1)$, 
$|\mathbf{f}|$ denotes the radial frequency in 2D Fourier space, and 
$f_\alpha \geq 0$ controls the spectral slope~\cite{harrington2025s}. 
The filter $1/(1{+}|\mathbf{f}|)^{f_\alpha}$ attenuates 
high-frequency components: $f_\alpha{=}0$ recovers white noise 
(uniform power spectrum), while larger $f_\alpha$ progressively 
suppresses fine-scale variation in favour of spatially coherent 
low-frequency structure. This shaping matters because 
high-frequency noise injected into transformer features tends to 
manifest as texture-level artefacts that the model treats as 
corruption~\cite{harrington2025s}, whereas low-frequency noise 
induces coherent spatial perturbations (e.g., layout shifts, 
compositional rearrangements) that align with the structural 
decisions distilled few-step models make in their early forward 
passes.

The noise is patchified identically to the features 
(Eq.~\ref{eq:patchify}), yielding $\boldsymbol{\epsilon}_\text{patches}$, and projected onto the PCA basis with 
singular-value weighting:
\begin{equation}
    \mathbf{c} = \boldsymbol{\epsilon}_\text{patches} \, \mathbf{V},
    \qquad
    \mathbf{c}_\text{scaled} = \mathbf{c} \odot
    \frac{\mathbf{s}}{\bar{s}},
    \qquad
    \mathbf{d}_\text{patches} = \mathbf{c}_\text{scaled} \,
    \mathbf{V}^\top
    \label{eq:pca_projection}
\end{equation}
where $\mathbf{V} \in \mathbb{R}^{P^2 D \times K}$ is the PCA basis 
from Eq.~\ref{eq:pca}, $\mathbf{s} = (s_1, \ldots, s_K)$ are the 
corresponding singular values, and $\bar{s}$ is their mean. The 
projection $\mathbf{c} = \boldsymbol{\epsilon}_\text{patches} \mathbf{V}$ 
expresses the patchified pink noise in the basis of the model's 
dominant variance directions, discarding the orthogonal complement 
that lies off the local feature manifold. The singular-value 
weighting allocates perturbation energy according to the model's 
intrinsic variance structure, ensuring that dominant modes (e.g., 
global structure and layout) are emphasized while suppressing 
directions corresponding to noise or unstable features. The resulting 
directed perturbation $\mathbf{d}_\text{patches}$ is then unpatchified 
back to the full spatial grid and injected via 
Eq.~\ref{eq:stride_injection}, ensuring that all perturbations remain 
aligned with the model's learned feature geometry.

\section{Experiments}
\label{sec:experiments}
 
\subsection{Experimental Setup}
\label{sec:setup}
 
\noindent \textbf{Models.}
We evaluate on two primary models spanning single-step and few-step inference \citep{sauer2024adversarial, sauer2024fast, luo2023latent, song2023consistency, esser2024scaling, liu2023instaflow, chen2025sana}:
\textbf{FLUX.1-schnell}~\cite{flux2024}, a 12B-parameter MMDiT with 1-step distilled inference (19 double-stream blocks, 38 single-stream blocks, 4096 tokens at $1024{\times}1024$), and \textbf{SD3.5 Large Turbo}~\cite{sd35}, an 8.1B-parameter MMDiT distilled to 4 denoising steps(38 joint transformer blocks, 4096 tokens at $1024{\times}1024$). These models represent the two dominant acceleration paradigms: single-step (no iterative refinement) and few-step distillation (compressed denoising schedule).

\noindent \textbf{Datasets.}
We evaluate on four standard text-to-image benchmarks: \textbf{MS-COCO}~\cite{lin2014coco} for large-scale distributional diversity, \textbf{DrawBench}~\cite{saharia2022photorealistic} for compositionally challenging prompts, \textbf{PartiPrompts}~\cite{yu2022parti} for broad category coverage, and \textbf{GenEval}~\cite{ghosh2023geneval} for compositional accuracy. For MS-COCO, we use a subset of 2{,}000 captions from the 2014 validation split. See Supp. Sec~B.1.

\begin{table*}[t]
\centering
\caption{%
  \textbf{Results on DrawBench} (199 prompts $\times$ 4 images) and
  \textbf{PartiPrompts} (1632 prompts $\times$ 4 images).
  Among single-pass training-free methods, STRIDE  achieves 
  the best diversity-quality tradeoff on both backbones, 
  Pareto-dominating all baselines on FLUX. On SD3.5 Turbo, CADS 
  collapses$^{\dagger}$. Per-prompt standard deviations and 
  significance analysis in Tables~11 and 12. \textbf{Bold} indicates the better result between STRIDE and its no-PCA ablation.%
}
\label{tab:drawbench_parti}
\small
\setlength{\tabcolsep}{4pt}
\renewcommand{\arraystretch}{1.2}
\begin{subtable}{\textwidth}
\centering
\caption{DrawBench (199 prompts $\times$ 4 images)}
\begin{tabular}{l cccc cccc}
\toprule
& \multicolumn{4}{c}{\textbf{FLUX.1-schnell}~1-Step} & \multicolumn{4}{c}{\textbf{SD3.5 Turbo}~4-Step} \\
\cmidrule(lr){2-5}\cmidrule(lr){6-9}
\textbf{Method}
  & \textbf{InBSim}$\downarrow$ & \textbf{VD/p}$\uparrow$ & \textbf{CLIP}$\uparrow$ & \textbf{HPS}$\uparrow$
  & \textbf{InBSim}$\downarrow$ & \textbf{VD/p}$\uparrow$ & \textbf{CLIP}$\uparrow$ & \textbf{HPS}$\uparrow$ \\
\midrule
Baseline           & 0.662          & 2.236          & 28.47          & 0.293          & 0.707                     & 2.067                     & 28.18                    & 0.291 \\
CADS               & 0.556          & 2.578          & 27.09          & 0.290          & 0.318$^\dagger$           & 3.309$^\dagger$           & 22.67$^\dagger$          & 0.262$^\dagger$ \\
Input Noise        & 0.566          & 2.570          & 27.43          & 0.284          & 0.668                     & 2.203                     & 28.34                    & 0.292 \\
DDC                & ---            & ---            & ---            & ---            & 0.600                     & 2.436                     & 27.84                    & 0.274 \\
SPELL              & ---            & ---            & ---            & ---            & 0.638                     & 2.334                     & 26.32                    & 0.274 \\
\midrule
STRIDE No-PCA          & 0.673          & 2.205          & \textbf{29.04}          & 0.280          & 0.704                     & 2.082                     & 28.07                    & \textbf{0.291} \\
\rowcolor{ourgrey}
\textbf{STRIDE (Ours)} & \textbf{0.590} & \textbf{2.496} & 28.61 & \textbf{0.284} & \textbf{0.639}            & \textbf{2.300}            & \textbf{28.07}           & 0.288 \\
\bottomrule
\end{tabular}
\end{subtable}

\vspace{8pt}

\begin{subtable}{\textwidth}
\centering
\caption{PartiPrompts (1632 prompts $\times$ 4 images)}
\begin{tabular}{l cccc cccc}
\toprule
& \multicolumn{4}{c}{\textbf{FLUX.1-schnell}~1-Step} & \multicolumn{4}{c}{\textbf{SD3.5 Turbo}~4-Step} \\
\cmidrule(lr){2-5}\cmidrule(lr){6-9}
\textbf{Method}
  & \textbf{InBSim}$\downarrow$ & \textbf{VD/p}$\uparrow$ & \textbf{CLIP}$\uparrow$ & \textbf{HPS}$\uparrow$
  & \textbf{InBSim}$\downarrow$ & \textbf{VD/p}$\uparrow$ & \textbf{CLIP}$\uparrow$ & \textbf{HPS}$\uparrow$ \\
\midrule
Baseline           & 0.722          & 2.025          & 28.13          & 0.290          & 0.769                     & 1.855                     & 28.09                    & 0.289 \\
CADS               & 0.630          & 2.328          & 27.37          & 0.288          & 0.356$^\dagger$           & 3.179$^\dagger$           & 22.88$^\dagger$          & 0.257$^\dagger$ \\
Input Noise        & 0.632          & 2.345          & 27.31          & 0.283          & 0.740                     & 1.959                     & 28.02                    & 0.288 \\
DDC                & ---            & ---            & ---            & ---            & 0.675                     & 2.187                     & 27.70                    & 0.273 \\
SPELL              & ---            & ---            & ---            & ---            & 0.690                     & 2.157                     & 26.69                    & 0.269 \\
\midrule
STRIDE No-PCA          & 0.744          & 1.952          & \textbf{28.77}          & 0.273          & 0.766                     & 1.870                     & \textbf{28.00}                    & \textbf{0.287} \\
\rowcolor{ourgrey}
\textbf{STRIDE (Ours)} & \textbf{0.653} & \textbf{2.275} & 28.56 & \textbf{0.281} & \textbf{0.714}            & \textbf{2.051}            & 27.91           & 0.286 \\
\bottomrule
\end{tabular}
\end{subtable}
\end{table*}
 
\noindent \textbf{Metrics.}
We report four primary metrics. \textbf{InBatchSim 
(InBSim)}~$\downarrow$ measures the average pairwise CLIP 
similarity among images generated from the same prompt, 
directly quantifying mode collapse, where lower values indicate 
greater diversity. \textbf{Vendi Score~\citep{friedman2022vendi} 
per prompt (VD/p\textsubscript{DINO})}~$\uparrow$ computes the 
effective number of distinct outputs per prompt using 
DINO~\citep{oquab2023dinov2} features, providing a feature-level 
diversity measure complementary to InBSim. \textbf{CLIP 
Score}~\citep{hessel2021clipscore}~$\uparrow$ measures 
text-image alignment, ensuring diversity gains do not come at 
the expense of prompt faithfulness. \textbf{Human Preference 
Score (HPSv2)}~\citep{wu2023hps}~$\uparrow$ evaluates 
perceptual quality and aesthetic appeal; we refer to HPSv2 
simply as \textbf{HPS} throughout the remainder of the paper 
for brevity. For COCO dataset, we additionally report 
\textbf{KID}~\citep{binkowski2018demystifying}~$\downarrow$ 
to measure distributional distance to real images. For 
GenEval, we additionally report \textbf{DreamSim 
(DrSim)}~\citep{fu2023dreamsim}~$\uparrow$, the average pairwise 
distance among generations from the same prompt, 
to align with the diversity-evaluation protocol of 
\cite{harrington2025s}, which we compare against on this 
benchmark.
 
\noindent \textbf{Baselines.}
We compare STRIDE against five training-free diversity methods: \begin{enumerate}[leftmargin=*,nosep] 
\item \textbf{CADS}~\cite{sadat2023cads}: Condition-Annealed Diffusion Sampling, which adds noise to text conditioning embeddings. For single-step models (FLUX), the annealing schedule collapses to a single noise level; for SD3.5 Turbo (4 steps), the full
per-timestep schedule operates as designed. 
\item \textbf{Input Noise}: Pink latent noise blended into the initial Gaussian noise before denoising, following the frequency-shaped approach of Harrington~et~al.~\cite{harrington2025s}: $\mathbf{z}' = (1{-}\alpha)\,\mathbf{z} +\alpha\,\mathbf{z}_\text{pink}$.
\item \textbf{STRIDE No-PCA} (ablation): Pink noise added directly to transformer features \textit{without} spatial PCA projection, serving as the ``without on-manifold projection'' ablation of our method.
\item \textbf{DDC}~\cite{gandikota2026distilling}: We use the Hybrid Inference variant of \cite{gandikota2026distilling} (Algorithm 1, $k{=}1$): SD3.5 Large (teacher) for the first denoising step, SD3.5 Turbo (student) for steps 2--4.
\item \textbf{SPELL}~\cite{kirchhof2024shielded}: Applied to SD3.5 Turbo only (structurally inapplicable to single-step FLUX); both hyperparameters $(r, \lambda)$ are jointly swept following the paper's calibration protocol (Appendix~B.2).
\end{enumerate}
All baseline hyperparameters are selected via grid search to identify their respective Pareto-optimal operating points; see Supp. Sec.~B.2 for grids and final values.
Statistical robustness is verified across both prompts (per-prompt standard errors, Table~11) and seeds (three-seed reproducibility, Table~12).

\noindent \textbf{STRIDE.}
STRIDE  is applied to the \textit{early} transformer blocks at 
the \textit{first} denoising step only, consistent with our finding 
(Supp.\ Sec.C) and the analysis 
of \cite{gandikota2026distilling} that compositional structure 
is established at step 0 in the early MM-DiT layers. Two design 
choices adapt to each model: (i)~the layer set spans roughly the 
first third of the dual-stream blocks (L0--7 of 19 for FLUX, 
L0--11 of 38 for SD3.5 Turbo); (ii)~we scale the perturbation 
magnitude $\alpha$ inversely with the number of denoising steps 
to prevent compounding across the schedule, with diversity 
primarily modulated through the frequency-shaping exponent 
$f_\alpha$. We use $K{=}1024$ principal components for both backbones\footnote{We also tested $K{=}256$ and observed comparable Pareto behaviour after re-tuning $\alpha$.}. Per-model hyper-parameters are reported in 
Supp.\ Sec.~B.3.

\begin{table*}[t]
\centering
\caption{%
  \textbf{Results on GenEval} (553 prompts $\times$ 4 images; 
  FLUX at $512^2$, SD3.5-Turbo at $1024^2$). Among single-pass 
  training-free methods, STRIDE achieves the best 
  diversity-quality tradeoff on both backbones, improving 
  diversity while preserving CLIP and HPS. On SD3.5 Turbo, 
  CADS collapses$^{\dagger}$. Rows marked $^{\star}$ are taken 
  directly from~\cite{harrington2025s} for a comparable 
  FLUX-schnell setting. Per-prompt standard deviations in 
  Table~11; further setup details in 
  Sec.~B.2. \textbf{Bold} indicates the better result between STRIDE and its no-PCA ablation.%
}
\label{tab:geneval}
\small
\setlength{\tabcolsep}{3pt}
\renewcommand{\arraystretch}{1.15}
\begin{tabular}{@{} l ccccc ccccc @{}}
\toprule
& \multicolumn{5}{c}{\textbf{FLUX.1-schnell}~1-Step} & \multicolumn{5}{c}{\textbf{SD3.5 Turbo}~4-Step} \\
\cmidrule(lr){2-6}\cmidrule(lr){7-11}
\textbf{Method}
  & \textbf{InBSim}$\downarrow$ & \textbf{DrSim}$\uparrow$ & \textbf{VD/p}$\uparrow$ & \textbf{CLIP}$\uparrow$ & \textbf{HPS}$\uparrow$
  & \textbf{InBSim}$\downarrow$ & \textbf{DrSim}$\uparrow$ & \textbf{VD/p}$\uparrow$ & \textbf{CLIP}$\uparrow$ & \textbf{HPS}$\uparrow$ \\
\midrule
Baseline           & 0.669          & 0.363          & 2.204          & 28.34          & 0.296          
                   & 0.741          & 0.264          & 1.948          & 29.20 & 0.290          \\
CADS               & 0.559 & 0.431          & 2.550          & 27.34          & 0.291          
                   & 0.299$^\dagger$ & 0.634$^\dagger$ & 3.348$^\dagger$ & 22.82$^\dagger$ & 0.257$^\dagger$ \\
Input Noise        & 0.592          & 0.445          & 2.476          & 27.32          & 0.287          
                   & 0.701          & 0.314          & 2.087          & 28.92          & 0.291 \\
DDC                & --            & --            & --            & --            & --            
                   & 0.627          & 0.368            & 2.343          & 28.52          & 0.274          \\
SPELL                & --            & --            & --            & --            & --            
                   & 0.651          & 0.369           & 2.284          & 27.88          & 0.272          \\
\midrule
i.i.d.$^{\star}$~\cite{harrington2025s}     & --     & 0.307          & 2.013          & --             & 0.304 
                                            & --     & --             & --             & --             & --             \\
GI$^{\star}$~\cite{parmar2025scaling}       & --     & 0.413          & 2.473          & --             & 0.296          
                                            & --     & --             & --             & --             & --             \\
NoiseOpt$^{\star}$~\cite{harrington2025s}   & --     & 0.446 & 2.753 & --             & 0.293          
                                            & --     & --             & --             & --             & --             \\
\midrule
STRIDE No-PCA          & 0.696          & 0.322          & 2.123          & \textbf{29.26} & 0.279          
                   & 0.730          & 0.274          & 1.986          & \textbf{29.13}          & \textbf{0.290}          \\
\rowcolor{ourgrey}
\textbf{STRIDE (Ours)} &\textbf{ 0.607}          & \textbf{0.423}          & \textbf{2.428}          & 29.04          & \textbf{0.282}
                   & \textbf{0.679}          & \textbf{0.337}          & \textbf{2.175}          & 28.86          & 0.287          \\
\bottomrule
\end{tabular}
\end{table*}

\subsection{Main Results}
\label{sec:main_results}

\begin{figure*}[t]
    \centering
    \begin{subfigure}{0.49\textwidth}
        \centering
        \includegraphics[width=\textwidth]{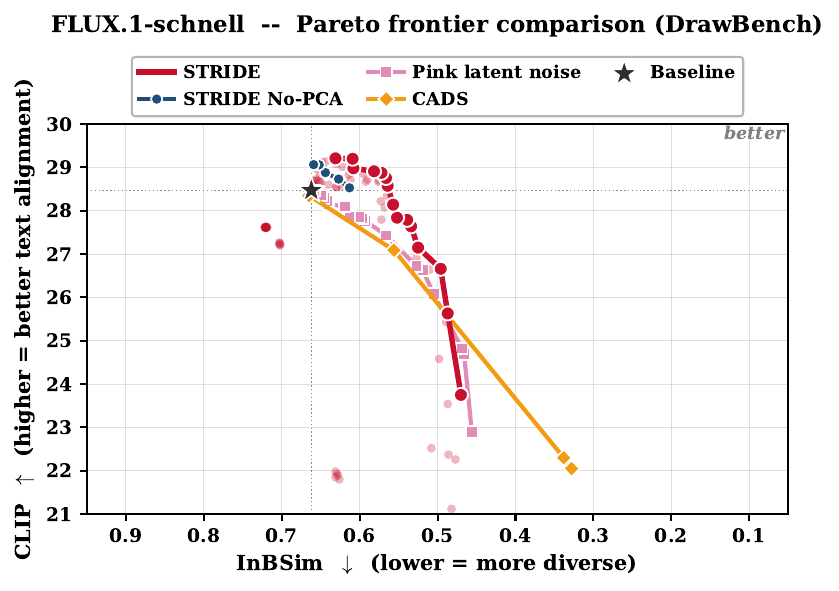}
        \caption{FLUX.1-schnell}
        \label{fig:flux_pareto_main}
    \end{subfigure}\hfill
    \begin{subfigure}{0.49\textwidth}
        \centering
        \includegraphics[width=\textwidth]{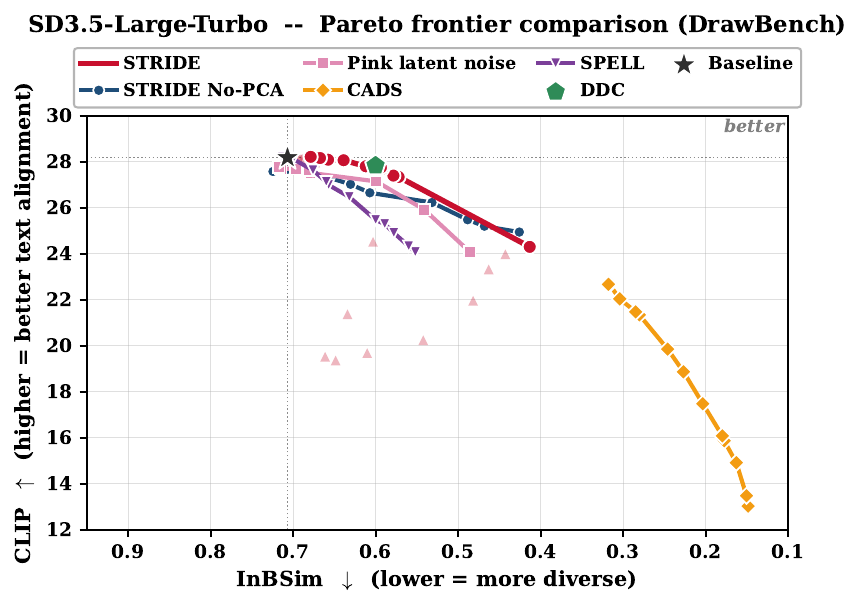}
        \caption{SD3.5 Turbo}
        \label{fig:sd35_pareto_main}
    \end{subfigure}
    \caption{\textbf{Pareto frontiers for diversity-quality trade-off.} We compare STRIDE against baselines using InBSim$\downarrow$ and CLIP$\uparrow$ on FLUX.1-schnell and SD3.5 Turbo. STRIDE achieves the strongest Pareto frontier on both architectures, improving diversity while preserving image-text alignment.}
    \label{fig:pareto_main}
\end{figure*}

\begin{figure*}[t]
    \centering
    \begin{subfigure}{0.49\textwidth}
        \centering
        \includegraphics[width=\textwidth]{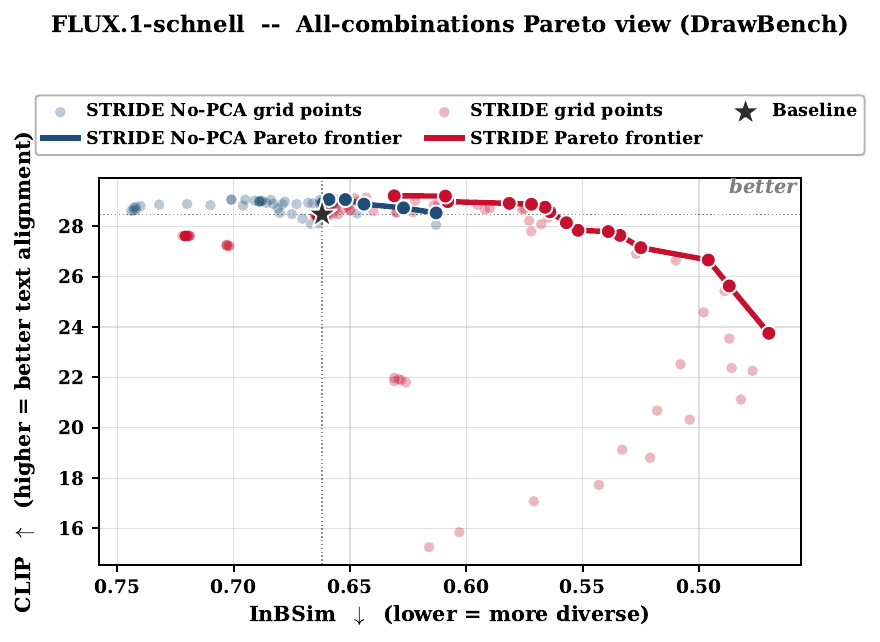}
        \caption{FLUX.1-schnell}
        \label{fig:flux_pareto_secondary}
    \end{subfigure}\hfill
    \begin{subfigure}{0.49\textwidth}
        \centering
        \includegraphics[width=\textwidth]{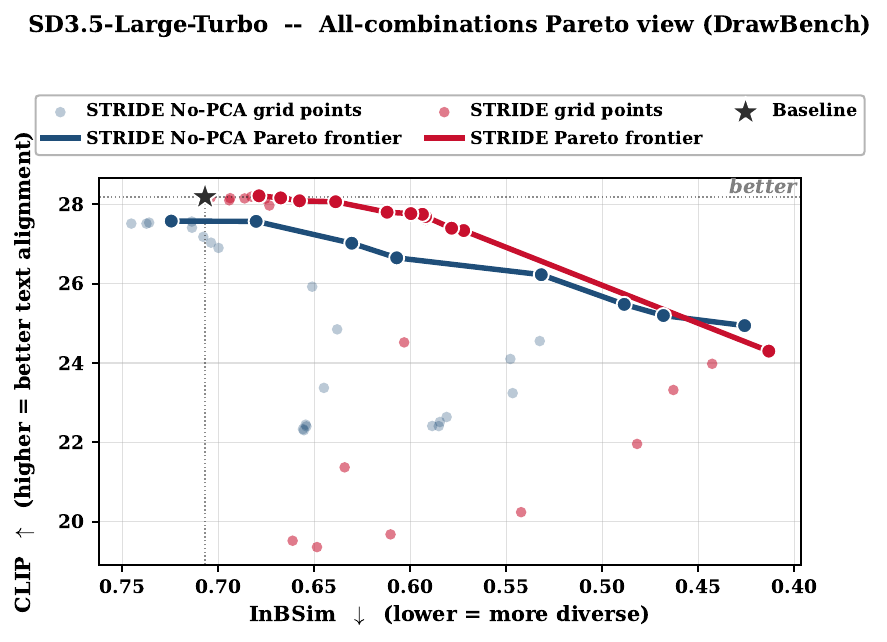}
        \caption{SD3.5 Turbo}
        \label{fig:sd35_pareto_secondary}
    \end{subfigure}
    \caption{\textbf{Diversity-quality Pareto frontier on DrawBench.} We compare STRIDE No-PCA (undirected pink noise) and STRIDE (PCA-directed pink noise) across frequency exponent $f_\alpha$ and perturbation strength $\alpha$ on FLUX.1-schnell and SD3.5 Turbo using 199 prompts with 4 images per prompt. STRIDE offers a better diversity-quality trade-off, achieving lower InBSim at comparable CLIP across both backbones.}
    \label{fig:pareto_secondary}
\end{figure*}

\noindent \textbf{FLUX.1-schnell.}
Table~\ref{tab:main_coco} presents results on COCO, and Table~\ref{tab:drawbench_parti}
on DrawBench and PartiPrompts.
STRIDE is the \textit{only} method that simultaneously improves both diversity and
text alignment across all four datasets.
On COCO, STRIDE achieves a $-7.5\%$ reduction in InBSim ($0.683$ vs.\ baseline $0.738$)
while \textit{improving} CLIP by $+0.96$ ($27.66$ vs.\ $26.69$).
On DrawBench, the diversity gain is $-10.9\%$ InBSim with CLIP improving by $+0.14$.
On PartiPrompts, $-9.6\%$ InBSim with $+0.43$ CLIP.
By contrast, both CADS and input noise achieve comparable or greater diversity
reductions but consistently degrade CLIP ($-0.25$ to $-1.38$ across datasets).
This establishes STRIDE as Pareto-dominant: at any given diversity level,
it achieves higher text alignment than all baselines, as visualized in Figures~\ref{fig:flux_pareto_main} and~\ref{fig:flux_pareto_secondary}; qualitative comparisons against the baseline are shown in Figure~\ref{fig:qualitative_teaser}.
On GenEval (Table~\ref{tab:geneval}), STRIDE reaches DreamSim $0.423$ and Vendi/p $2.428$, comparable to the concurrent test-time noise-optimization method of \cite{harrington2025s} ($0.446$, $2.753$) 
while requiring \textit{no} optimization loop and adding less inference cost (single forward pass vs.\ ${\sim}80$ backward iterations per prompt).

\noindent \textbf{SD3.5 Large Turbo.}
The advantage of STRIDE is even more pronounced on SD3.5 Large Turbo, where CADS collapses entirely (CLIP$<$23 at every tested configuration;
Table~\ref{tab:main_coco}). Input noise provides only moderate diversity ($-2.9\%$ InBSim on COCO, $-5.5\%$ on DrawBench), while STRIDE No-PCA barely moves ($-0.7\%$ on COCO).
STRIDE achieves $-6.0\%$ InBSim on COCO and $-9.6\%$ on DrawBench with CLIP essentially preserved ($-0.07$ and $-0.12$ respectively), and again attains the second lowest KID ($8.68$ vs.\ baseline $9.62$). STRIDE on SD3.5 Turbo achieves meaningful diversity improvement without quality collapse (Figures~\ref{fig:sd35_pareto_main} and~\ref{fig:sd35_pareto_secondary}). DDC-Hybrid~\cite{gandikota2026distilling} reaches a comparable Pareto point (InBSim $0.60$, CLIP $27.84$), but requires loading 
both the SD3.5 Large teacher and the distilled student at inference: $56.16$\,GB co-resident VRAM versus STRIDE's $28.11$\,GB, a $2{\times}$ memory cost that exceeds a single A100-40GB and forces CPU offload, raising wall-clock to 
$10.5$\,s/image ($+805\%$ vs.\ baseline). STRIDE attains the same trade-off using only the distilled student in a single forward pass at $1.50$\,s/image ($+29\%$ at $K{=}256$). 
SPELL~\cite{kirchhof2024shielded} matches STRIDE's diversity on DrawBench ($-9.8\%$ InBSim) but at $15{\times}$ the CLIP cost ($-1.86$ vs.\ STRIDE's $-0.12$), reflecting the steep quality penalty of repulsion-based methods on few-step schedules.

\noindent \textbf{Hyperparameter ablation.} Patch size $P{=}2$ is uniquely effective: increasing $P$ to $\{4, 8, 16\}$ reverts InBSim toward baseline on both backbones (Table~\ref{tab:p_ablation_main}), because the joint patch dimension $P^2 D$ grows quadratically, leaving the projected subspace too narrow to register. We adopt $P{=}2$ as the locked configuration. For full ablation see Supp. Sec C and Tables~15 and~21; per-prompt standard 
deviations and significance analysis are reported in Supp. Sec~B.5; qualitative comparisons across configurations are in Supp. Sec~D.

\noindent \textbf{Human Evaluation.}
We conduct a two-alternative forced-choice (2AFC) study on 24
randomly sampled DrawBench prompts with \textbf{119} participants (50 completed all 24 questions, 69 partial).
For each prompt, evaluators are shown a grid of 4 baseline images 
and 4 STRIDE images in randomized order and asked to select the 
set exhibiting greater visual diversity. Across $1{,}370$ total responses, STRIDE is
preferred \textbf{67.9\%} of the time ($930/1{,}370$; 95\% Wilson
CI: $[65.4\%, 70.3\%]$; binomial test $p < 10^{-39}$; Table~10). STRIDE 
wins 22 of 24 prompts (18 at $p < 0.05$), with one tie; the 
single exception is a prompt requesting a specific fictional 
character, where baseline outputs are already highly constrained 
and evaluators favour the more canonical depiction (see Supp.\ 
Sec.~B.4 for full details). For 
detailed analysis and ablation, see Supp.\ 
Sec. C.

\section{Observations}
\label{sec:observation}
 
\noindent \textbf{Why PCA is essential.} STRIDE No-PCA (unstructured pink noise injected directly at the feature level, without PCA projection)
shows that on-manifold projection is \textit{necessary}, not merely beneficial. On FLUX, unstructured noise makes outputs \textit{more} similar: COCO InBSim \textit{rises} $+3.3\%$ ($0.762$ vs.\ baseline $0.738$) and PartiPrompts $+3.0\%$ ($0.744$ vs.\ $0.722$). However, PCA projection inverts this 
at matched perturbation magnitude, yielding $-7.5\%$ InBSim on COCO. The model's internal correction mechanisms respond to off-manifold perturbation by converging toward its mode, tightening the output distribution. PCA-directed perturbation overcomes this by aligning noise with the model's learned variance structure.
 
\noindent \textbf{Layer selection and timestep gating.}
On both MM-DiT models, diversity originates from perturbation at 
\textbf{early} transformer blocks (L0--7 for FLUX, L0--11 for 
SD3.5 Turbo); late blocks produce negligible diversity, consistent 
with the interpretation that early blocks set spatial composition 
while later blocks refine local 
detail~\cite{park2025softpag,jiang2025no}. For SD3.5 Turbo's 4-step schedule, perturbation at \textbf{step~0} 
is most effective; early steps commit to the global composition, so late-step perturbations can only alter local details (see Supp.\ Table~20).

\begin{table}[t]
\centering
\caption{%
  \textbf{Patch size $P$ ablation on DrawBench} (199 prompts 
  $\times$ 4 images) at per-backbone 
  perturbation strength ($\alpha{=}0.5$ for FLUX, $\alpha{=}0.1$ for SD3.5 Turbo). The locked configuration 
  $P{=}2$ (bolded); larger $P$ values dilute per-patch PCA coverage, reducing diversity gains.
}
\label{tab:p_ablation_main}
\small
\setlength{\tabcolsep}{4pt}
\renewcommand{\arraystretch}{1.15}
\begin{tabular}{l c cccc c cccc}
\toprule
& & \multicolumn{4}{c}{\textbf{FLUX.1-schnell} ($\alpha{=}0.5$)} & & \multicolumn{4}{c}{\textbf{SD3.5 Turbo} ($\alpha{=}0.1$)} \\
\cmidrule(lr){3-6}\cmidrule(lr){8-11}
\textbf{Metric} & \textbf{Base} & \boldmath$P{=}2$ & $P{=}4$ & $P{=}8$ & $P{=}16$ & \textbf{Base} & \boldmath$P{=}2$ & $P{=}4$ & $P{=}8$ & $P{=}16$ \\
\midrule
InBSim$\downarrow$ & 0.664 & \textbf{0.590} & 0.663 & 0.663 & 0.664 & 0.741 & \textbf{0.639} & 0.693 & 0.701 & 0.701 \\
CLIP$\uparrow$     & 28.40 & \textbf{28.61} & 28.50 & 28.39 & 28.41 & 29.20 & \textbf{28.07} & 28.18 & 28.18 & 28.16 \\
\bottomrule
\end{tabular}
\end{table}

\section{Conclusion}
\label{sec:conclusion}

We present \textbf{STRIDE}, a training- and optimization-free method for improving output diversity in distilled few-step diffusion models through PCA-directed spatial feature perturbation. 
Our central finding is that the 
perturbation geometry matters more than magnitude: 
identical noise energy reduces InBSim by $7.5\%$ when projected onto the model's principal components, but increases it by $3.3\%$ when applied unstructured (COCO, 2{,}000 prompts Table~\ref{tab:main_coco}).
STRIDE operates via a single forward hook with no iterative optimization. Across FLUX.1-schnell (1-step) and SD3.5 Large Turbo (4-step), STRIDE Pareto-dominates all training-free baselines on the diversity-fidelity frontier. On FLUX, it is the only method that simultaneously improves diversity \emph{and} text alignment; on SD3.5 Large Turbo, it is the only single-forward-pass method to achieve meaningful diversity, while CADS collapses entirely. Alternative paradigms based on noise optimization~\cite{harrington2025s}, post-hoc selection~\cite{parmar2025scaling}, or hybrid base-distilled inference~\cite{gandikota2026distilling} can also improve diversity, but require iterative search, overgeneration, or access to the undistilled model; STRIDE is complementary to these approaches.

\noindent \textbf{Limitations.}
STRIDE has three practical costs.  \textbf{(i) Residual artefacts.} 
Because perturbation occurs inside the forward pass, single-step 
models have no subsequent denoising steps to attenuate residual 
noise; generated images may retain subtle high-frequency 
artefacts at large $\alpha$ or $f_\alpha$, unlike multi-step 
models where later refinement steps smooth the perturbation. 
\textbf{(ii) Compute overhead.} Online PCA adds wall-clock 
overhead that scales with $K$, and prompt-dependence of the 
spatial principal directions 
(Supp. Table~6) precludes a simple 
precomputed basis. More discussion in Supp. Sec C 
\textbf{(iii) Architecture-specific tuning.} 
The optimal layer set and timestep gating differ across model 
families (L0--7 for FLUX, L0--11 with step-0-only timestep gating for 
SD3.5 Turbo); transferring STRIDE to a new architecture 
requires re-identifying the early structure-forming blocks.

\noindent \textbf{Future work.}
Three directions follow naturally:  (1)~Amortised PCA to reduce per-image overhead; (2)~Composing STRIDE with 
conditioning-level and latent-level diversity methods to test whether gains are super-additive; and (3)~Automating noise injection-site selection for new architectures.

\bibliographystyle{plain}
\bibliography{main}

\end{document}